\documentclass[runningheads]{llncs}
\usepackage{multirow}  

\usepackage{amsmath}   
\usepackage{amssymb}   
\usepackage{bm}        
\usepackage{enumerate} 
\usepackage{graphicx}  
\usepackage{caption}
\usepackage{graphicx}
\usepackage{subfigure}
\usepackage{setspace}
\usepackage{verbatim}  
\begin{document}
\title{End-to-End Pareto Set Prediction with Graph Neural Networks for Multi-objective Facility Location}
\titlerunning{End-to-End Pareto Set Prediction with Graph Neural Networks}
%
\author{Shiqing Liu\inst{1} \and
Xueming Yan\inst{*2,1} \and
Yaochu Jin\inst{*1}}
\authorrunning{S. Liu et al.}
%
\institute{Faculty of Technology, Bielefeld University, 33619 Bielefeld, Germany \and 
School of Information Science and Technology, Guangdong University of Foreign Studies, Guangzhou 510000, China \\
\email{yanxm@gdufs.edu.cn;yaochu.jin@uni-bielefeld.de}
}
\maketitle              
%
\begin{abstract}
    The facility location problems (FLPs) are a typical class of NP-hard combinatorial optimization problems, which are widely seen in the supply chain and logistics. Many mathematical and heuristic algorithms have been developed for optimizing the FLP. In addition to the transportation cost, there are usually multiple conflicting objectives in realistic applications. It is therefore desirable to design algorithms that find a set of Pareto solutions efficiently without enormous search cost. In this paper, we consider the multi-objective facility location problem (MO-FLP) that simultaneously minimizes the overall cost and maximizes the system reliability. We develop a learning-based approach to predicting the distribution probability of the entire Pareto set for a given problem. To this end, the MO-FLP is modeled as a bipartite graph optimization problem and two graph neural networks are constructed to learn the implicit graph representation on nodes and edges. The network outputs are then converted into the probability distribution of the Pareto set, from which a set of non-dominated solutions can  be sampled non-autoregressively. 
    Experimental results on MO-FLP instances of different scales show that the proposed approach achieves a comparable performance to a widely used multi-objective evolutionary algorithm in terms of the solution quality while significantly reducing the computational cost for search.

\end{abstract}
\vspace{-0.5cm}
\section{Introduction}
\label{sect:introduction}
Multi-objective combinatorial optimization (MOCO) has received considerable attention over the past few decades due to its wide applications in the real-world. In MOCO, there are multiple conflicting objectives, and it is often non-trivial to optimize them simultaneously \cite{bengio2021machine}. The multi-objective facility location problem (MO-FLP) is a typical NP-hard MOCO problem. It aims to determine an optimal set of facility locations that can satisfy all the customer demands within certain constraints, while minimizing the total costs and maximizing the system reliability. Decisions made in facility location have a long-term impact on numerous operational and logistical strategies and are critical to both private and public firms \cite{hale2003location}.

A lot of work has been devoted to developing mathematical methods or handcrafted heuristic algorithms for solving MOCO problems. 
An intuitive approach is to reduce a multi-objective problem to a single-objective problem by calculating the weighted sum of multiple objectives. However, assigning a suitable weight to each objective introduces an additional hyperparameter optimization problem. 
Evolutionary algorithms (EAs) have been successful in approximating the Pareto set of MOCOs by maintaining and updating a set of solutions \cite{deb2002fast,zhang2007moea,cheng2016reference}. However, EAs and other population-based methods often require a large number of function evaluations during the search process, incurring prohibitive computing overhead when the objective functions are expensive to evaluate \cite{jin2005comprehensive}. Moreover, it is difficult to resue the knowledge about the optimal sets of solutions for other instances of the same problem that have already been solved.

Most existing work considers MOCO as constrained mixed-integer linear programming, overlooking the highly structured nature of the combinatorial optimisation problems. 
For example in the facility location problem (FLP), the locations of all facilities and customers can be represented by a set of nodes separately, and the transport overhead is the weight of the edge connecting two nodes from different sets. Generally, permutation-based COPs can be formulated as sequential decision-making tasks on graphs \cite{joshi2019efficient}, and matching-based COPs can be considered as node and edge classification or prediction tasks on graphs. Therefore, machine learning methods can be used to extract high-dimensional characteristics of the graph-based problems and learn optimal policies to solve COPs instead of relying on handcrafted heuristics \cite{bengio2021machine,vesselinova2020learning}. Graph neural networks (GNNs) can exploit the message passing scheme to learn the structural information of nodes and edges efficiently according to the graph topology. Consequently, GNNs are well-suited for tackling the MOCO problems \cite{hudson2021graph,gasse2019exact,cappart2021combinatorial}. However, most existing methods focus on solving permutation-based problems and only consider one single objective, neglecting the study of more commonly seen matching-based multi-objective COPs \cite{farahani2019or}. 

\begin{figure}[tb]
	\begin{centering}
    \setlength{\abovecaptionskip}{0.cm}
	\includegraphics[width=1.0\textwidth]{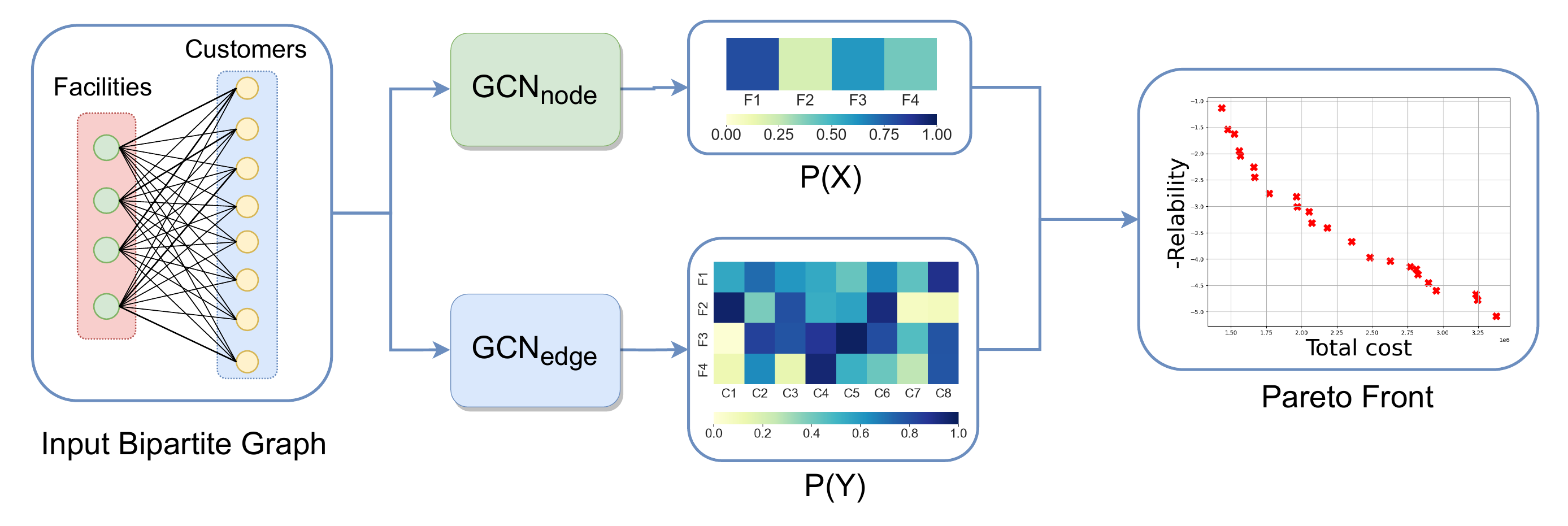}
    \caption{An overview of the proposed approach. The MO-FLP instance is converted into a bipartite graph as the input to the GCNs. The two GCN models perform node classification and edge prediction and output the probability models, which are co-sampled to generate a set of non-dominated solutions in a one-shot manner.}
	\label{fig:framework}
	\end{centering}
\end{figure}

In this paper, we propose a learning-based approach leveraging graph convolutional networks (GCNs) to approximate the Pareto set distribution of the multi-objective facility location problem. The overall framework is shown in Fig.~\ref{fig:framework}. The problem is formulated as a bipartite graph with edge connections between two independent sets of nodes. The model consists of two different residual gated GCNs for node classification and edge prediction tasks, respectively. The model takes bipartite graphs as the input, and transforms the original node and edge features into high-dimensional embeddings. Several residual gated graph convolutional layers are used to learn the structural information from the graph topology and update the embeddings iteratively. The output of the first GCN is a prediction of the probability of each factory being selected in the Pareto optimal solutions. The output of the second GCN is a probabilistic model in the form of an adjacency matrix, denoting the probability of each customer being assigned to each selected factory. The output probability models can be sampled directly to generate a set of Pareto solutions in a one-shot manner. The two networks are trained coordinately by supervised learning. The training data is a large set of MO-FLP instances with various Pareto optimal solutions generated by a multi-objective evolutionary algorithms, e.g., the fast elitism non-dominated sorting genetic algorithm (NSGA-II) \cite{deb2002fast}. The main contributions of this paper include:
\vspace{-0.1cm}
\begin{enumerate}
    \item We formulate the MO-FLP as a bipartite graph optimization task and develop a novel learning-based combinatorial optimization method to directly approximate the Pareto set of new instances of the same problem without extra search.
    \item We propose an end-to-end probabilistic prediction model based on two GCNs for node and edge predictions, respectively, and train the model with a supervised learning using data generated by a multi-objective evolutionary algorithm.
    \item We demonstrate the efficiency of our proposed method for solving MO-FLP instances with different scales. Our experimental results show that the learning-based approach can approximate a set of Pareto optimal solutions without additional search, significantly reducing the computational cost compared to population-based algorithms.
\end{enumerate}

\section{Related Work}
\label{sect:relatedwork}
\vspace{-0.1cm}

\subsection{Facility Location Problem}
FLPs are a typical class of NP-hard combinatorial problems in operations research \cite{laporte2019introduction}. FLPs consider choosing an optimal set of facilities among all the potential sites and determines an allocation scheme for all customers, under the constraints that all customer demands must be satisfied by the constructed facilities. A common objective of FLPs is to minimize the total costs, which consist of the transportation cost and the fixed cost. 

FLPs have several variants depending on different constraint settings and objective functions. Each candidate facility may have a limited or unlimited maximum capacity, which classifies the problems into capacitated and uncapacitated facility location problems. When the number of established facilities is fixed to $k$, there are two variants, namely the $k$-median problem \cite{vasilyev2019effective} and the $k$-center problem \cite{chakrabarty2020non}. The $k$-median problem minimizes the sum of distances from each customer to the closest facility, while the $k$-center problem minimizes the maximum value of a distance from a customer to the closest facility. Another category of variants is the covering problem, where the problems share a property that a customer can receive the service only if it is located with a certain distance from the nearest facility \cite{garcia2015covering}. The set covering problem aims to find a set of facilities with the minimum number that can satisfy all customers' demands. The maximum covering problem intends to find a set of facilities with a fixed number to maximize the total demands it covers. From an objective perspective, FLPs and its variants can be divided into single- and multi-objective problems. In addition to the overall costs, multi-objective facility location problems may also include other practical objectives such as the system reliability in logistics, which is quite desirable in real-world applications \cite{farahani2019or}.

\vspace{-0.2cm}
\subsection{Graph Representation Learning}

Graph-structured data is ubiquitous in daily life. Various kinds of data can be naturally expressed as graphs, such as social relationships, telecommunication networks, chemical molecules, and also combinatorial optimization tasks \cite{wu2020comprehensive}. Generally, a graph is a collection of objects (nodes) along with a set of interactions (edges) between pairs of them \cite{hamilton2020graph}. With the development of machine learning techniques, graph representation learning has attracted increasing attention for in-depth analysis and effective utilization of graph data. Graph representation learning derives node and edge embeddings based on the graph topology for a variety of downstream tasks in machine learning, such as node classification \cite{wang2020nodeaug}, edge prediction \cite{kim2019edge}, and graph clustering. The traditional graph representation methods neither use the node features nor share parameters in the encoder, and are not able to generalize to unseen nodes after training.
To alleviate these limitations, graph neural networks are proposed to learn node embeddings in a more explainable way based on the topology and attributes of the input graph \cite{wu2020comprehensive,zhou2020graph}. Early attempts made by Sperduti and Starita \cite{sperduti1997supervised} dealt with arbitrary structured data as directed acyclic graphs with recursive neural networks. Gori \cite{gori2005new} and Scarselli \cite{scarselli2008graph} generalized the recursive neural networks for other types of graph structures and introduced the concept of graph neural networks. With the compelling performance shown by convolutional neural networks in computer vision tasks, a lot of work has been devoted to the transfer of convolution operators to graph domain \cite{zhang2019graph}, which can be categorized into spectral-based methods \cite{bruna2013spectral,defferrard2016convolutional,levie2018cayleynets} and spatial-based methods \cite{gilmer2017neural,niepert2016learning,monti2017geometric}. GNNs have been practically applied to various domains and achieved encouraging performance \cite{joshi2019efficient,hudson2021graph,gasse2019exact}.

\vspace{-0.2cm}
\subsection{Machine Learning for Combinatorial Optimization on Graphs}
\vspace{-0.1cm}
NP-hard combinatorial optimization problems are non-trivial to solve, but the instances are relatively easy to generate. In many practical scenarios, decision-makers need to solve different instances of the same optimization task, where the instances share the same problem structure and only differ in data \cite{bengio2021machine,khalil2017learning,cappart2021combinatorial}. Traditional heuristic methods require extensive expert knowledge and a huge computational cost, and their solutions cannot be transferred to other instances. To address this limitation, recent years have seen a surge in research on machine learning approaches to combinatorial optimization to automate the solution of different instances of combinatorial optimization problems \cite{vesselinova2020learning,cappart2021combinatorial}. Combinatorial optimization problems often depict a collection of entities and their relations, which are graph-structured data in essence. Therefore, many GNN-based machine learning methods are proposed to solve combinatorial optimization problems \cite{joshi2019efficient,joshi2019learning,khalil2017learning}. 
Kool et al. developed a GNN model in an encoder-decoder architecture based on attention layers, and trained it using REINFORCE for solving routing problems \cite{kool2019attention}. In addition to solving COPs directly, machine learning techniques can also be used to provide valuable information to operation research algorithms \cite{gasse2019exact}. 
Although a lot of effort has been devoted to developing ML methods for COPs, most work has focused on single-objective permutation-based problems, and little research on multi-objective matching-based problems has been reported. 

\vspace{-0.2cm}
\section{Problem Formulation}
\vspace{-0.2cm}
\label{sect:problemformulation}
This section begins with a formal definition of the multi-objective uncapacitated facility location problem, which is mathematically formulated as integer linear programming. Subsequently, we discuss how to measure the logistics system reliability in facility location problems.

\textbf{Multi-objective uncapacitated facility location.} 
Consider a set of candidate facility locations and a set of demand points (customers) with fixed locations. Every customer has its own quantity of demand to be satisfied. Each potential facility has its own fixed cost for construction, and there are different transportation costs between facilities and customers associated with their distances. Each customer should be served by only one facility, while each facility can serve multiple customers simultaneously. The target is to identify the selected collection of facilities for construction and assign an allocation plan for all customers, in order to minimize the total costs and maximize the system reliability.

\textbf{Mathematical Formulation.} 
With the minimization of the total costs (including fixed costs and transportation costs) and the maximization of the system reliability as the two objectives, the multi-objective uncapacitated facility location problem can be defined as follows:

\begin{equation}
    \min C_{\text {total}}=\sum_{i \in M} f_i X_i+\sum_{i \in M} \sum_{j \in N} q_j d_{i j} c_{i j} Y_{i j}
    \label{equ:minC}
\end{equation}

\begin{equation}
    \max R_{\text {sys}}=\frac{\sum_{i \in M} \sum_{j \in N} q_j X_i\left[1-F_{V_{i j}}\left(\frac{d_{i j}}{t_j}\right)\right]}{\sum_{j \in N} q_j}
    \label{equ:maxR}
\end{equation}

\begin{equation}
    \text {s.t.  } \quad \sum_{i \in M} Y_{i j}=1 \quad j \in N 
\end{equation}

\begin{equation}
    X_i , Y_{i j}=\{0,1\}, \; Y_{i j} \leq X_i \quad i \in M, j \in N
\end{equation}

In the MO-FLP, assume there are $m$ candidate facility locations denoted as a set $M=\{1,2, \ldots, m\}$, and $n$ customer points denoted as a set $N=\{1,2, \ldots, n\}$. $f_{i} \in \mathbb{R^{+}}$ denotes the fixed cost of constructing the facility at candidate location $i \left (  i \in M \right )$, and $q_{j} \in \mathbb{R^{+}}$ is the demand volume of customer $j \left (  j \in N \right )$. $d_{i j} \in \mathbb{R^{+}}$ and $c_{i j} \in \mathbb{R^{+}}$ are the distance and the unit transportation cost between facility $i$ and customer $j$ respectively. $V_{i j}$ denotes the speed for vehicles travelling from facility $i$ to customer $j$, and $F_{V_{i j}}( \cdot )$ is a statistically regular velocity distribution. $t_{j}$ is the delivery timescale required by customer $j$.

The decision variables are $X_i \in \{0,1\}$, which denotes whether facility location $j$ is selected ($X_i=1$) or not ($X_i=0$), and $Y_{i j} \in \{0,1\}$, which denotes whether customer $j$ is served by facility $i$ ($Y_{i j}=1$) or not ($Y_{i j}=0$). The two objectives are the minimization of the total costs $C_{\text {total}}$ and maximization of the system reliability $R_{\text {sys}}$.

\textbf{Logistics system reliability.} 
Reliability is the probability that a system performs its intended function under the stated conditions \cite{rausand2003system}. Logistics system reliability is defined as the probability at which the system will successfully provide services to customers under certain conditions and within a specified time. System reliability is a common metric for assessing service levels in modern logistics. The service reliability $R_{ij}$ between factory $i$ and customer $j$ is defined as:

\begin{equation}
    R_{i j}=P\left(T_{i j} \leq t_j\right)=P\left(\frac{d_{i j}}{V_{i j}} \leq t_j\right)=P\left(V_{i j} \geq \frac{d_{i j}}{t_j}\right)=1-F_{V_{i j}}\left(\frac{d_{i j}}{t_j}\right),
\end{equation}
where $T_{i j}$ is the time cost for delivery from facility $i$ to customer $j$, and $F_{V_{i j}}( \cdot )$ is a statistically regular velocity distribution function that usually follows the characteristics of a normal distribution. Based on this, the logistics system reliability of facilities serving multiple customers is calculated by:
\begin{equation}
    \setlength{\abovedisplayskip}{0pt}
    \setlength{\belowdisplayskip}{0pt}
    R_{\mathrm{sys}}=\frac{\sum_{i \in M} \sum_{j \in N} q_j R_{i j}}{\sum_{j \in N} q_j}=\frac{\sum_{i \in M} \sum_{j \in N} q_j\left[1-F_{V_{i j}}\left(\frac{d_{i j}}{t_j}\right)\right]}{\sum_{j \in N} q_j}
\end{equation}

\vspace{-0.2cm}
\section{Method}
\label{sect:method}
\vspace{-0.2cm}
We first convert an MO-FLP instance to a bipartite graph based on its inherent structural properties, and then train a dual GCN-based model to directly output the probabilistic model of the Pareto optimal solutions for the given task. The proposed model consists of two graph convolutional networks $GCN_{node}$ and $GCN_{edge}$. $GCN_{node}$ learns high-dimensional representations of nodes and outputs a probabilistic prediction for each node via a simple multi-layer perceptron (MLP) classifier. Meanwhile, $GCN_{edge}$ learns high-dimensional edge representations and predicts the probability of each edge appearing in the Pareto optimal solutions in the form of an adjacency matrix. The entire model is trained in an end-to-end manner by minimizing the loss between predictions and ground-truth labels. During the test, the output probabilistic model is sampled and converted into a set of non-dominated solutions in a non-autoregressive manner, eliminating the requirement of further search when solving new instances.

\vspace{-0.4cm}
\subsection{Bipartite Optimization on MO-FLP}
An instance of the MO-FLP is transformed into a bipartite graph $G=(U, V, E)$, whose vertices are divided into two independent sets $U$ (including all candidate facilities) and $V$ (including all customers), and these two parts are connected by a set of edges $E$. Within the graph, each facility in $U$ contains the information of its fixed cost, while each customer in $V$ contains its demand and delivery timescale information. The features of each edge in $E$ contain the Euclidean distance, the transportation cost and the reliability between the facility and the customer it connects. The aim of converting the MO-FLP into a bipartite optimization is to derive high-dimensional embeddings in the latent space through graph representation learning, in order to predict optimal solutions by means of machine learning.


\vspace{-0.2cm}
\subsection{The Dual GCN-based model}

\textbf{Overall framework.} Note that a solution to an MO-FLP problem consists of two parts: $ X = \{ X_i \mid i \in M \}$ and $ Y = \{ Y_{ij} \mid i \in M, j \in N \}$. The decision variable $X$ first determines a subset of facilities to be constructed from all candidate locations, then the decision variable $Y$ identifies the allocation scheme between customers and the selected locations in $X$. According to the mathematical formulation in Section~\ref{sect:problemformulation}, the calculation of objective $C_{total}$ in Equation~\ref{equ:minC} requires both $X$ and $Y$ as the decision variables, while the second objective $R_{sys}$ in Equation~\ref{equ:maxR} is only determined by $X$. Leveraging the structural properties of the MO-FLP problem discussed above, we propose to  predict the two components $X$ and $Y$ by designing two GCN models, one for node prediction and the other for edge prediction.

As shown in Fig.~\ref{fig:framework}, the proposed model consists of $GCN_{node}$ and $GCN_{edge}$, which take the same bipartite graph as their input. 
More specifically, $GCN_{node}$ loads node and edge information and computes $H$-dimensional representations for each node via iterative graph convolution operators. 
The last graph convolution layer is followed by a multi-layer perceptron (MLP) classifier, where the updated node embeddings are taken as its inputs to compute the probability of each node being selected in decision variable $X$. The output of the classifier is represented as a probabilistic model $P(X) \in \mathbb{R}^{M}$, where $M$ is the number of all candidate facilities. Simultaneously, $GCN_{edge}$ takes the same node and edge information as input attributes and derives $H$-dimensional representations for each edge. A following edge classifier is used to predict the probability of each edge occurring in the Pareto optimal solutions in the form of a heat-map over the adjacency matrix $P(Y) \in \mathbb{R}^{M \times N}$, where $N$ is the number of all customers. The outputs of the two GCN models indicate the information of $X$ and $Y$, respectively, which together constitute a prediction of the Pareto optimal solutions. 
The GCN architectures adopted in the proposed model consist of three building blocks: an embedding block, a graph convolution block and an MLP classifier. 

\textbf{Embedding block.} The inputs to the embedding block are a set of original node features $\mathbf{h_{n}}=\left\{\vec{u}_1, \vec{u}_2, \ldots, \vec{u}_M, \vec{v}_1, \vec{v}_2, \ldots, \vec{v}_N\right\}, \vec{u}_i \in \mathbb{R}^{F_{u}}, \vec{v}_j \in \mathbb{R}^{F_{v}}$ and edge features $\mathbf{h_{e}}=\left\{\vec{w}_{11}, \vec{w}_{12}, \ldots, \vec{w}_{MN}\right\}, \vec{w}_{ij} \in \mathbb{R}^{F_{e}}$. $M$ and $N$ are the numbers of facilities and customers, and $F_{u}, F_{v}$ and $F_{e}$ are the numbers of features for different nodes and edges. The outputs of the embedding block are node embeddings $\mathbf{n}=\left\{\vec{n}_{1}, \vec{n}_{2}, \ldots, \vec{n}_{M+N}\right\}, \vec{n}_{i} \in \mathbb{R}^{H}$ and edge embeddings $\mathbf{e}=\left\{\vec{e}_{11}, \vec{e}_{12}, \ldots, \vec{e}_{MN}\right\}, \vec{e}_{ij} \in \mathbb{R}^{H}$, where $H$ is the dimension of the hidden space.

For node embeddings, each feature $a \in \mathbb{R}$ is first embedded in a $d$-dimensional vector $\vec{\alpha} \in \mathbb{R}^{d}$ by a learnable linear transformation to get adequate expressive power. Then all the feature vectors are concatenated together to get an embedding $\vec{n}_i$ for node $i$:

\begin{equation}
    \setlength{\abovedisplayskip}{3pt}
    \setlength{\belowdisplayskip}{3pt}
    \vec{n}_i=\operatorname{concat}_{k=1}^{F_n}\left(\vec{\alpha}_i^k\right)
\end{equation}

Similarly, the edge embedding $\vec{e}_{ij}$\textbf{} for the edge between node $i$ and node $j$ is the concatenation of all the edge feature vectors:

\begin{equation}
    \vec{e}_{ij}=\operatorname{concat}_{k=1}^{F_e}\left(\vec{\beta}_{ij}^k\right)
\end{equation}

The selection of node and edge features as the input to the embedding layers depends on the problem's characteristics, which should have a significant impact on the objective function values. For the MO-FLP problem investigated in this work, there are several candidate node features of the bipartite graph served as input: the node category of the binary classification (i.e., whether a node belongs to the facility set or the customer set), the demand volume of a customer, the fixed cost of constructing a facility, the transportation costs and the reliability of all edges connected to a node. And the input edge features include the adjacency matrix of the bipartite graph, the transportation cost, and the reliability of an edge. 

\textbf{Graph convolution block.} The message passing process mainly occurs in the graph convolution block by stacking several graph convolution layers sequentially. It leverages the structure and properties of the input graph in order to exchange information between neighbors and update node and edge embeddings without changing the connectivity. The graph convolution adopted in our model follows the framework of residual gated graph convolutional neural network \cite{bresson2017residual}, where additional edge features and residual gated operators are integrated to introduce heterogeneity in the message passing process. 


In the graph convolution block, the inputs to the $k$-th layer are a set of node embeddings $\mathbf{n^{k}}=\left\{\vec{n}_{1}^{k}, \vec{n}_{2}^{k}, \ldots, \vec{n}_{M+N}^{k}\right\}$ and a set of edge embeddings $\mathbf{e^{k}}=\left\{\vec{e}_{11}^{k}, \vec{e}_{12}^{k}, \ldots, \vec{e}_{MN}^{k}\right\}$ where $\vec{n}_i, \vec{e}_{ij} \in \mathbb{R}^{H}$. The $k$-th layer outputs an update set of both node and edge embeddings with the same dimension $H$.

Let $\vec{e}_{ij}^{k}$ denote the edge embedding between node $i$ and node $j$ at the $k$-th GCN layer. In the message passing of $\vec{e}_{ij}$ (the superscript $k$ is omitted for simplicity), we first gather the associated node embeddings $\vec{n}_{i}$ and $\vec{n}_{j}$ as neighborhood information, and aggregate all the messages as $\vec{e}_{i j}^{\,\prime}$. Then $\vec{e}_{i j}^{\,\prime}$ is passed through a batch normalization layer $\operatorname{BN}$ and the rectified linear unit $\operatorname{ReLU}$, to form the updated edge embedding $\vec{e}_{ij}^{k+1}$ together with the original input $\vec{e}_{ij}^{k}$: :

\begin{equation}
    \setlength{\abovedisplayskip}{5pt}
    \setlength{\belowdisplayskip}{5pt}
    \vec{e}_{i j}^{k+1}=\vec{e}_{i j}^{k}+\operatorname{ReLU}\left(\operatorname{BN}\left(\textbf{U} \vec{e}_{ij}^{k} + \textbf{V} \left(\vec{n}_{i}^{k} + \vec{n}_{j}^{k}\right)\right)\right),
\end{equation}
where $\textbf{U}, \textbf{V} \in \mathbb{R}^{H \times H}$ are linear transformations. Suppose $\vec{n}_{i}^{k}$ denotes the node embedding of node $i$ at the $k$-th layer. For updating $\vec{n}_{i}$, we first calculate the weight vector $\bm{\omega}_{ij}$ of each neighbor node $j$ as:

\begin{equation}
    \setlength{\abovedisplayskip}{5pt}
    \setlength{\belowdisplayskip}{5pt}
    \bm{\omega}_{i j}=\frac{\sigma\left(\vec{e}_{i j}\right)}{\sum_{j \in \mathcal{N}_i} \sigma\left(\vec{e}_{i j}\right)+\delta},
\end{equation}
where $\mathcal{N}_i$ denotes all the first-order neighbors of node $i$. $\sigma$ represents the sigmoid function, and $\delta > 0$ is a small value. Then we gather the neighbor embeddings $\vec{n}_{j} \, (j \in \mathcal{N}_i)$ and define the output of the $k$-th convolution layer as:

\begin{equation}
    \vec{n}_{i}^{k+1}=\vec{n}_{i}^{k}+\operatorname{ReLU}\big(\operatorname{BN}\big(\textbf{P}\vec{n}_i+ \textbf{Q} \sum_{j \in \mathcal{N}_i} \bm{\omega}_{i j} \vec{n}_j \big)\big),
\end{equation}
where $\textbf{P}, \textbf{Q} \in \mathbb{R}^{H \times H}$ are linear transformations. The stack of graph convolution layers enables neighborhood messages to be progressively transferred within the graph. The dimensionality of the embeddings remains the same, however, the representation of each node and edge contains more local information in addition to its original features.

\textbf{MLP classifier.} The updated representations are taken as inputs to an MLP for classification tasks. For node prediction in $GCN_{node}$, we consider $\vec{n}_i \, (i \in M)$ as the high-dimensional embedding of node $i$ from the facility set $M$. For edge prediction in $GCN_{edge}$, we consider $\vec{e}_{ij} \, (i \in M, j \in N)$ as the embedding of edge between facility $i$ and customer $j$. The probability $\hat{p}_i \in \left[ 0,1 \right]$ of node $i$ being selected as a constructed facility and the probability $\hat {p}_{ij} \in \left[ 0,1 \right]$ of facility $i$ serving customer $j$ are predicted by:

\begin{equation}
    \setlength{\abovedisplayskip}{5pt}
    \setlength{\belowdisplayskip}{5pt}
    \hat{p}_{i} = \operatorname{MLP}(\vec{n}_{i}), \; \hat{p}_{ij} = \operatorname{MLP}(\vec{e}_{ij})
\end{equation}

The weight parameters is trained in an end-to-end manner by minimizing the mean square error between the prediction $\hat{P}(X) = \left\{\hat{p}_i \mid i \in M \right\}$ and the ground-truth label $P(X) = \left\{p_i \mid i \in M \right\}$ via gradient descent methods.

Since each customer must be served by only one facility, we consider the edge prediction for each customer as a multi-class classification task and train the network parameters by minimizing the cross entropy loss between the prediction $\hat{P}(Y) = \left\{\hat{p}_{ij} \mid i \in M, j \in N \right\}$ and the ground-truth label $P(Y) = \left\{p_{ij} \mid i \in M, j \in N \right\}$, where $P(X)$ and $P(Y)$ are both derived from the Pareto optimal solutions.

\textbf{End-to-end training.} The dataset for training and testing the proposed model is generated by a multi-objective evolutionary algorithm. We generate MO-FLP instances of different scales (i.e., various numbers of facility and customer nodes) and approximate their Pareto Fronts via the fast elitist non-dominated sorting genetic algorithm (NSGA-II) \cite{deb2002fast}. Then the probabilistic distributions $P(X)$ and $P(Y)$ for each instance are derived from a set of Pareto optimal solutions, which serve as ground-truth labels for training and evaluating the proposed model.
\section{Experiments}
\vspace{-0.1cm}
\subsection{Dataset Generation and Hyperparameter Configurations}
\label{subsec:hyperparameter}
We consider MO-FLP problems with the following four different configurations: $M \times N$ are set to $20 \times 50$, $20 \times 100$, $50 \times 100$, and $50 \times 200$. We randomly generate 1000 instances for each problem scale and optimize them using NSGA-II until convergence to approximate the true Pareto fronts. Then the 1000 instances for each scale are divided into a training dataset, a validation dataset and a test dataset with 700, 200 and 100 pairs of instances and  ground-truth labels, respectively. During each training epoch, the training data is split into mini-batches with a batch size $B = 20$ instances. The Adam optimizer is used to train the weights of the proposed model with an initial learning rate of $\gamma = 0.001$ and a maximum number of 300 epochs. 
Both $GCN_{node}$ and $GCN_{edge}$ consist of $l_{GCN} = 3$ graph convolutional layers and $l_{MLP} = 3$ classification layers. The dimension of the hidden space is set to $H = 128$ for node and edge embeddings. During the test, we sample 200 solutions from the output prediction for each instance and calculate the hypervolume (HV) and IGD value of the obtained non-dominated solution sets as the performance indicators.

\subsection{Experimental Results}
\vspace{-0.1cm}

There are two variants of our proposed model adopted in the experiments, named $Dual\_A$ and $Dual\_B$ with different input features. $Dual\_A$ takes the node category, the customer demand and the fixed cost of each facility as the original node features, while $Dual\_B$ also considers the transportation costs and the service reliability of all the edges associated with the node. Both architectures share the same edge features as inputs. To investigate the model performance on MO-FLP with various scales, we compare it to NSGA-II with different numbers of function evaluations (MFEs). We set the number of independent runs to 20 for the compared algorithm, and calculate the mean and standard deviation of HV and IGD values as the performance indicators. The population size is set to 100 for all experiments.

\begin{table}[htb]
\centering
\footnotesize
\caption{The percentages of test instances where two variants of the proposed model perform better than NSGA-II with different MFEs in terms of the two indicators.}
\label{tab:betterpercent}
\begin{tabular}{lccccccc}
\hline
                        &                          & MFEs & 10000 & 20000 & 30000 & 40000 & 50000 \\ \hline
\multirow{4}{*}{$20\times50$}  & \multirow{2}{*}{$Dual\_A$} & HV  & 100\% & 98\%  & 96\%  & 75\%  & 41\%  \\
                        &                          & IGD & 100\% & 90\%  & 63\%  & 31\%  & 20\%  \\ \cline{2-8} 
                        & \multirow{2}{*}{$Dual\_B$} & HV  & 100\% & 98\%  & 88\%  & 69\%  & 29\%  \\
                        &                          & IGD & 100\% & 86\%  & 55\%  & 27\%  & 8\%   \\ \hline
\multirow{4}{*}{$20\times100$} & \multirow{2}{*}{$Dual\_A$} & HV  & 100\% & 100\% & 94\%  & 76\%  & 51\%  \\
                        &                          & IGD & 100\% & 100\% & 86\%  & 57\%  & 45\%  \\ \cline{2-8} 
                        & \multirow{2}{*}{$Dual\_B$} & HV  & 100\% & 100\% & 100\% & 100\% & 96\%  \\
                        &                          & IGD & 100\% & 100\% & 100\% & 98\%  & 90\%  \\ \hline
\multirow{4}{*}{$50\times100$} & \multirow{2}{*}{$Dual\_A$} & HV  & 100\% & 100\% & 100\% & 73\%  & 33\%  \\
                        &                          & IGD & 100\% & 94\%  & 63\%  & 29\%  & 2\%   \\ \cline{2-8} 
                        & \multirow{2}{*}{$Dual\_B$} & HV  & 100\% & 100\% & 96\%  & 61\%  & 27\%  \\
                        &                          & IGD & 100\% & 96\%  & 53\%  & 12\%  & 0\%   \\ \hline
\multirow{4}{*}{$50\times200$} & \multirow{2}{*}{$Dual\_A$} & HV  & 100\% & 100\% & 100\% & 76\%  & 14\%  \\
                        &                          & IGD & 100\% & 88\%  & 43\%  & 8\%   & 0\%   \\ \cline{2-8} 
                        & \multirow{2}{*}{$Dual\_B$} & HV  & 100\% & 100\% & 100\% & 100\% & 90\%  \\
                        &                          & IGD & 100\% & 100\% & 98\%  & 86\%  & 69\%  \\ \hline
\end{tabular}
\end{table}

Table~\ref{tab:betterpercent} shows the performance of the proposed model compared to NSGA-II in different problem scales. We train the dual GCN-based models with different scales of the problem instances and evaluate them on test datasets. 
For each test case, 200 solutions are first sampled from the predicted probability distribution and evaluated by the objective functions to get a set of non-dominated solutions. Then we calculate the mean HV and IGD values. 
Finally, for HV and IGD values associated with each $\operatorname{MFE}$ configuration of NSGA-II, we count the percentage of the cases in which the proposed model performs better than NSGA-II out of the 100 test cases. The statistical results in Table~\ref{tab:betterpercent} indicate that for an unseen instance, by only sampling 200 solutions from the model, the performance of the sampled solution set is already better than NSGA-II with more than 10000 function evaluations.

\begin{figure}[htb]
\setlength{\leftskip}{-0.35cm}
    \setlength{\abovecaptionskip}{0.cm}
	\includegraphics[width=1.05\textwidth]{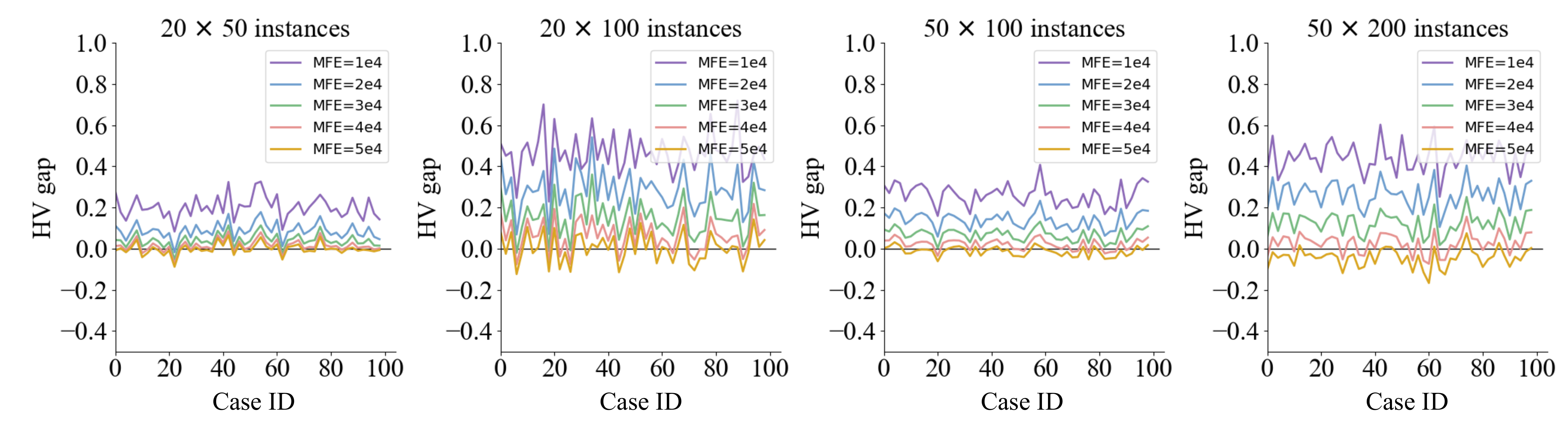}
	\caption{The difference in HV values between the proposed model and NSGA-II with different MFEs. 
 }
	\label{fig:lines}
\end{figure}

Figure~\ref{fig:lines} depicts the differences between the HV values of the solution sets obtained by NSGA-II and the proposed model for different problem scales with different MFEs. 
A positive difference means that the proposed model performs better than NSGA-II. These results reveal that the proposed model outperforms NSGA-II when the MFEs is less than 40000 in most test cases for all scales. In some cases the model performance is even comparable to that of NSGA-II with 50000 MFEs.

\vspace{-0.4cm}
\subsection{Hyperparameter Sensitive Analysis}
\begin{figure}[htbp]
    \setlength{\abovecaptionskip}{0cm}
    \setlength{\belowcaptionskip}{-0.5cm} 
    \subfigure[The impact of GCN layers.]{
        \includegraphics[width=0.48\linewidth]{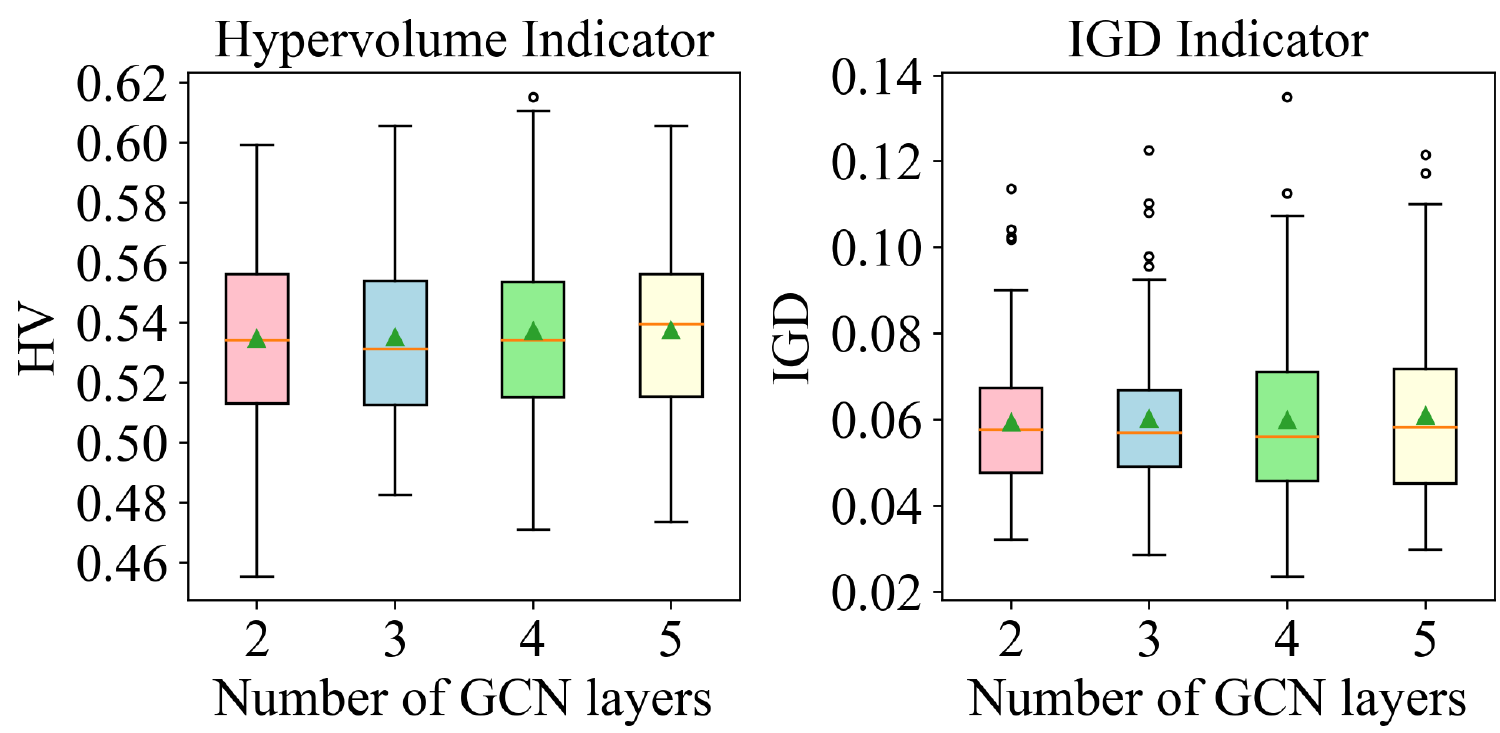}
        \label{fig:subfig1}
    }
    \subfigure[The impact of hidden dimensions.]{
	\includegraphics[width=0.48\linewidth]{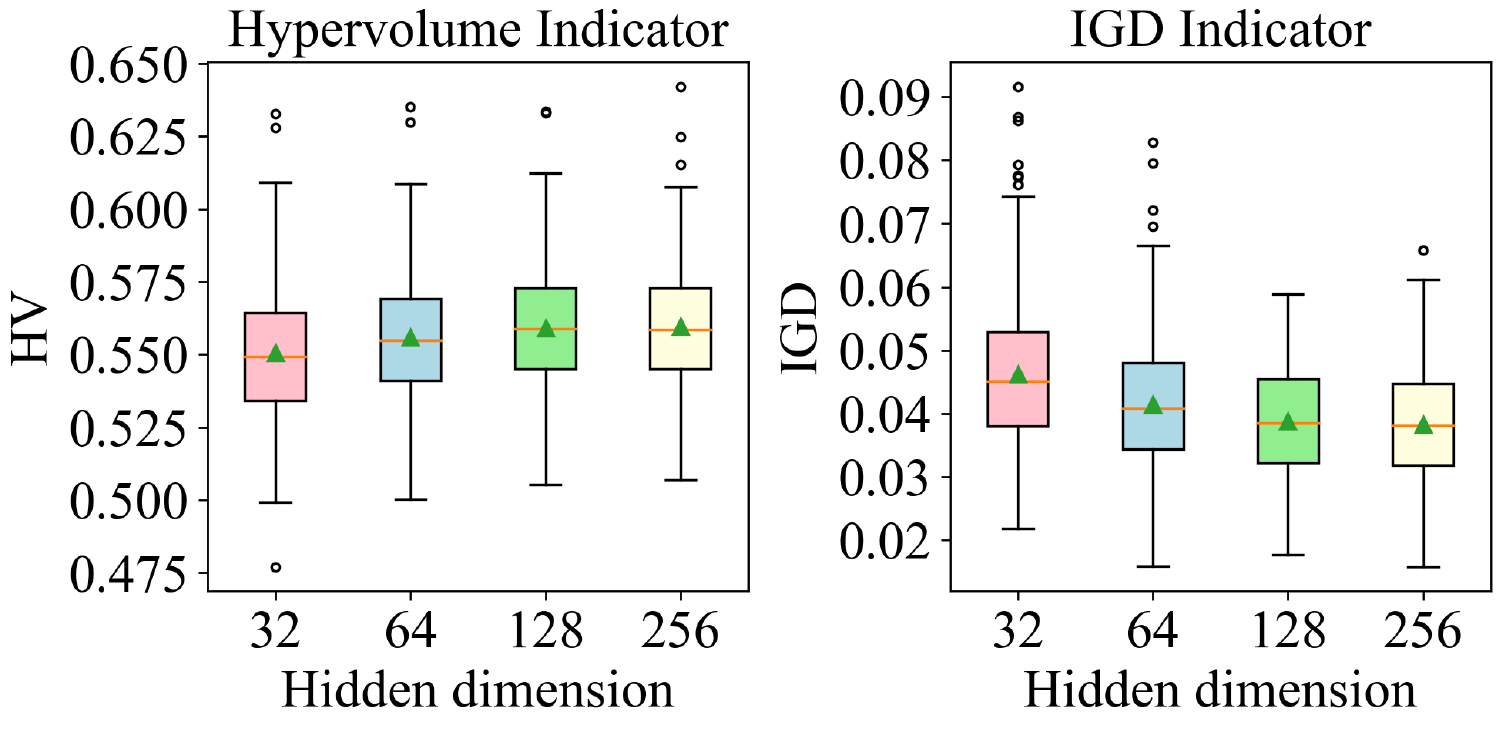}
	\label{fig:subfig2}
    }
    \caption{Sensitivity analysis. (a) The effect of different GCN layers. (b) The effect of different hidden dimensions.}
    
    \label{fig:hyperparameters}
\end{figure}

\vspace{-0.0cm}
We investigate the influence of different graph convolution layers and hidden dimensions on the two performance indicators, HV and IGD. We train the proposed model with different numbers of GCN layers on the $20 \times 20$ training dataset, and evaluate them on the test dataset with 100 unseen instances. The statistics of HV and IGD values are presented in the form of boxplots in Fig.~\ref{fig:subfig1}. The results demonstrate that the increase in the number of GCN layers has a little impact on the model performance, and $l_{GCN} = 3$ achieves a slightly better performance. Similarly, we train the proposed model for different dimensions of the hidden space and plot the statistical results of the two indicators in Fig.~\ref{fig:subfig2}. The model performance improves as the hidden dimension increases from 32 to 128. Note that a larger number of hidden layers and more GCN layers also lead to higher computational costs in the training process.

\vspace{-0.5cm}

\section{Conclusion and Future Work}
\vspace{-0.0cm}

This paper proposes a learning-based approach to directly predicting a set of non-dominated solutions for multi-objective facility location. We convert the original combinatorial optimization problem into a bipartite graph, and train two GCN models for predicting Pareto optimal solutions for unseen instances by learning the distribution of Pareto optimal solutions in previously solved examples. 
Experimental results on different scales of MO-FLP instances demonstrate that by only sampling hundreds of solutions, the proposed dual GCN-based approach can achieve a performance comparable to NSGA-II using up to tens of thousands of function evaluations. Future work will focus on improving the model scalability and exploring the heterogeneity of the input graphs in order to generalize the proposed approach to more complex and realistic problems with conflicting objectives and multiple constraints.

\subsubsection{Acknowledgment}
This work was supported in part by the National Natural Science Foundation of China under Grant No. 62006053, in part by the Program of Science and Technology of Guangzhou under Grant No. 202102020878 and in part by a Ulucu PhD studentship. Y. Jin is funded by an Alexander von Humboldt Professorship for Artificial Intelligence endowed by the German Federal Ministry of Education and Research.

%
%
%
%






\bibliographystyle{splncs04}
\bibliography{easychair}

\begin{thebibliography}{10}
\providecommand{\url}[1]{\texttt{#1}}
\providecommand{\urlprefix}{URL }
\providecommand{\doi}[1]{https://doi.org/#1}

\bibitem{bengio2021machine}
Bengio, Y., Lodi, A., Prouvost, A.: Machine learning for combinatorial
  optimization: a methodological tour d’horizon. European Journal of
  Operational Research  \textbf{290}(2),  405--421 (2021)

\bibitem{bresson2017residual}
Bresson, X., Laurent, T.: Residual gated graph convnets. arXiv preprint
  arXiv:1711.07553  (2017)

\bibitem{bruna2013spectral}
Bruna, J., Zaremba, W., Szlam, A., LeCun, Y.: Spectral networks and locally
  connected networks on graphs. In: Proceedings of the International Conference
  on Learning Representations (ICLR) (2014)

\bibitem{cappart2021combinatorial}
Cappart, Q., Ch{\'e}telat, D., Khalil, E., Lodi, A., Morris, C.,
  Veli{\v{c}}kovi{\'c}, P.: Combinatorial optimization and reasoning with graph
  neural networks. arXiv preprint arXiv:2102.09544  (2021)

\bibitem{chakrabarty2020non}
Chakrabarty, D., Goyal, P., Krishnaswamy, R.: The non-uniform k-center problem.
  ACM Transactions on Algorithms (TALG)  \textbf{16}(4),  1--19 (2020)

\bibitem{cheng2016reference}
Cheng, R., Jin, Y., Olhofer, M., Sendhoff, B.: A reference vector guided
  evolutionary algorithm for many-objective optimization. IEEE Transactions on
  Evolutionary Computation  \textbf{20}(5),  773--791 (2016)

\bibitem{deb2002fast}
Deb, K., Pratap, A., Agarwal, S., Meyarivan, T.: A fast and elitist
  multiobjective genetic algorithm: {NSGA-II}. IEEE Transactions on
  Evolutionary Computation  \textbf{6}(2),  182--197 (2002)

\bibitem{defferrard2016convolutional}
Defferrard, M., Bresson, X., Vandergheynst, P.: Convolutional neural networks
  on graphs with fast localized spectral filtering. Advances in Neural
  Information Processing Systems  \textbf{29} (2016)

\bibitem{farahani2019or}
Farahani, R.Z., Fallah, S., Ruiz, R., Hosseini, S., Asgari, N.: Or models in
  urban service facility location: A critical review of applications and future
  developments. European Journal of Operational Research  \textbf{276}(1),
  1--27 (2019)

\bibitem{garcia2015covering}
Garc{\'\i}a, S., Mar{\'\i}n, A.: Covering location problems. In: Location
  science, pp. 93--114. Springer (2015)

\bibitem{gasse2019exact}
Gasse, M., Ch{\'e}telat, D., Ferroni, N., Charlin, L., Lodi, A.: Exact
  combinatorial optimization with graph convolutional neural networks. Advances
  in Neural Information Processing Systems  \textbf{32} (2019)

\bibitem{gilmer2017neural}
Gilmer, J., Schoenholz, S.S., Riley, P.F., Vinyals, O., Dahl, G.E.: Neural
  message passing for quantum chemistry. In: Proceedings of the International
  Conference on Machine Learning. pp. 1263--1272. PMLR (2017)

\bibitem{gori2005new}
Gori, M., Monfardini, G., Scarselli, F.: A new model for learning in graph
  domains. In: Proceedings of the IEEE international Joint Conference on Neural
  Networks. vol.~2, pp. 729--734 (2005)

\bibitem{hale2003location}
Hale, T.S., Moberg, C.R.: Location science research: a review. Annals of
  Operations Research  \textbf{123}(1),  21--35 (2003)

\bibitem{hamilton2020graph}
Hamilton, W.L.: Graph representation learning. Synthesis Lectures on Artifical
  Intelligence and Machine Learning  \textbf{14}(3),  1--159 (2020)

\bibitem{hudson2021graph}
Hudson, B., Li, Q., Malencia, M., Prorok, A.: Graph neural network guided local
  search for the traveling salesperson problem. arXiv preprint arXiv:2110.05291
   (2021)

\bibitem{jin2005comprehensive}
Jin, Y.: A comprehensive survey of fitness approximation in evolutionary
  computation. Soft Computing  \textbf{9}(1),  3--12 (2005)

\bibitem{joshi2019efficient}
Joshi, C.K., Laurent, T., Bresson, X.: An efficient graph convolutional network
  technique for the travelling salesman problem. arXiv preprint
  arXiv:1906.01227  (2019)

\bibitem{joshi2019learning}
Joshi, C.K., Laurent, T., Bresson, X.: On learning paradigms for the travelling
  salesman problem. arXiv preprint arXiv:1910.07210  (2019)

\bibitem{khalil2017learning}
Khalil, E., Dai, H., Zhang, Y., Dilkina, B., Song, L.: Learning combinatorial
  optimization algorithms over graphs. Advances in Neural Information
  Processing Systems  \textbf{30} (2017)

\bibitem{kim2019edge}
Kim, J., Kim, T., Kim, S., Yoo, C.D.: Edge-labeling graph neural network for
  few-shot learning. In: Proceedings of the IEEE/CVF Conference on Computer
  Vision and Pattern Recognition. pp. 11--20 (2019)

\bibitem{kool2019attention}
Kool, W., Van~Hoof, H., Welling, M.: Attention, learn to solve routing
  problems! In: Proceedings of the International Conference on Learning
  Representations (ICLR) (2019)

\bibitem{laporte2019introduction}
Laporte, G., Nickel, S., Saldanha-da Gama, F.: Introduction to location
  science. In: Location science, pp. 1--21. Springer (2019)

\bibitem{levie2018cayleynets}
Levie, R., Monti, F., Bresson, X., Bronstein, M.M.: Cayleynets: Graph
  convolutional neural networks with complex rational spectral filters. IEEE
  Transactions on Signal Processing  \textbf{67}(1),  97--109 (2018)

\bibitem{monti2017geometric}
Monti, F., Boscaini, D., Masci, J., Rodola, E., Svoboda, J., Bronstein, M.M.:
  Geometric deep learning on graphs and manifolds using mixture model cnns. In:
  Proceedings of the IEEE Conference on Computer Vision and Pattern
  Recognition. pp. 5115--5124 (2017)

\bibitem{niepert2016learning}
Niepert, M., Ahmed, M., Kutzkov, K.: Learning convolutional neural networks for
  graphs. In: Proceedings of the International Conference on Machine Learning.
  pp. 2014--2023. PMLR (2016)

\bibitem{rausand2003system}
Rausand, M., Hoyland, A.: System reliability theory: models, statistical
  methods, and applications, vol.~396. John Wiley \& Sons (2003)

\bibitem{scarselli2008graph}
Scarselli, F., Gori, M., Tsoi, A.C., Hagenbuchner, M., Monfardini, G.: The
  graph neural network model. IEEE transactions on neural networks
  \textbf{20}(1),  61--80 (2008)

\bibitem{sperduti1997supervised}
Sperduti, A., Starita, A.: Supervised neural networks for the classification of
  structures. IEEE Transactions on Neural Networks  \textbf{8}(3),  714--735
  (1997)

\bibitem{vasilyev2019effective}
Vasilyev, I., Ushakov, A.V., Maltugueva, N., Sforza, A.: An effective heuristic
  for large-scale fault-tolerant k-median problem. Soft Computing
  \textbf{23}(9),  2959--2967 (2019)

\bibitem{vesselinova2020learning}
Vesselinova, N., Steinert, R., Perez-Ramirez, D.F., Boman, M.: Learning
  combinatorial optimization on graphs: A survey with applications to
  networking. IEEE Access  \textbf{8},  120388--120416 (2020)

\bibitem{wang2020nodeaug}
Wang, Y., Wang, W., Liang, Y., Cai, Y., Liu, J., Hooi, B.: Nodeaug:
  Semi-supervised node classification with data augmentation. In: Proceedings
  of the 26th ACM SIGKDD International Conference on Knowledge Discovery \&
  Data Mining. pp. 207--217 (2020)

\bibitem{wu2020comprehensive}
Wu, Z., Pan, S., Chen, F., Long, G., Zhang, C., Philip, S.Y.: A comprehensive
  survey on graph neural networks. IEEE Transactions on Neural Networks and
  Learning Systems  \textbf{32}(1),  4--24 (2020)

\bibitem{zhang2007moea}
Zhang, Q., Li, H.: Moea/d: A multiobjective evolutionary algorithm based on
  decomposition. IEEE Transactions on Evolutionary Computation  \textbf{11}(6),
   712--731 (2007)

\bibitem{zhang2019graph}
Zhang, S., Tong, H., Xu, J., Maciejewski, R.: Graph convolutional networks: a
  comprehensive review. Computational Social Networks  \textbf{6}(1),  1--23
  (2019)

\bibitem{zhou2020graph}
Zhou, J., Cui, G., Hu, S., Zhang, Z., Yang, C., Liu, Z., Wang, L., Li, C., Sun,
  M.: Graph neural networks: A review of methods and applications. AI Open
  \textbf{1},  57--81 (2020)

\end{thebibliography}

\end{document}